# Polyhedron Volume-Ratio-based Classification for Image Recognition

Qingxiang Feng, Jeng-Shyang Pan, Senior Member, IEEE,
Jar-Ferr Yang, Fellow, IEEE and Yang-Ting Chou

*Abstract*—In this paper, a novel method, called polyhedron volume ratio classification (PVRC), is proposed for image recognition. The linear regression based classification (LRC) and class mean (CM) approaches aim to find the closest projection to the subspace formed by the prototype data vectors in each class. Better than LRC and CM classifiers, the PVRC classifier computes the ratio of two polyhedron's volumes, where the first inclusive polyhedron is enclosed by the test plus class prototype data vectors while the second exclusive polyhedron is enclosed by class prototype data vectors. The PVRC tries to find the least ratio of class-based inclusive and exclusive polyhedron volumes among all classes. With mathematical optimization, the PVRC classifier achieves better recognition rate than the existing statistical methods for object and face recognition. With the proposed fast algorithm, it is noted that the computational cost of the proposed classifier is very small. A large number of experiments on Coil100 object database, Eth80 object database, Soil47 object database, GT face database and UMIST face database are used to evaluate the proposed algorithm. The experimental results demonstrate that the proposed method achieves better recognition rate and less computational cost than the existing methods, such as LRC, CM, nearest neighbor (NN), sparse representation based classification (SRC) and two phase test sample sparse representation (TPTSSR) classifiers.

## I. Introduction

FACE and object recognition systems are known to be critically dependent on classifiers. Nearest neighbor (NN) [1] and class mean (CM) [2] methods, which are designed to minimize the distance from the test vector to the subspace spanned by all training vectors of each object class, are the well-known approaches in pattern recognition area. The NN classifies the test sample based on the best representation in terms of a single training sample, while the CM classifies it based on the best mean representation in terms of all the training samples in each class.

Samples from a specific object class are known to lie on a linear subspace [3]-[4]. Borrowing the above subspace concept, the locally linear regression (LLR) [11] is proposed specifically to solve the problem of pose identification. Linear regression-based classification (LRC) [12], which is proposed for the problem of face identification, uses the similar concept to develop class-specific models of the registered users and renovate the task of face recognition as a problem of linear regression. For face recognition, the LRC approaches, including kernel LRC, Improved PCA-LRC, LDA-LRC and Unitary-LRC [7]-[10], have been proposed to further improve the recognition performance under different situations such as variable illumination and facial expression.

Instead of the class-model suggested in the LRC approaches, sparse representation based classification (SRC) [11-12] uses all-class-model to classify the test sample. Based on the SRC classifier, some improved methods are further presented for face recognition. These improved classifiers could be classified into two categories. The first category utilizes the novel representations of each class, in which all use the first phase to choose some prototype samples while the second phase is different [13]-[16]. Chang *et. al* utilizes the collaborative representation instead of the sparse representation [17]. The second category uses the sparse representation for the subspace learning/discriminant analysis, which employs the discriminant projection tensor discriminant projection [18], [19] or the sparse eigenface [20] to select the features. SRC classifier gains better performance for face recognition under variations of illumination, noise, and corruption. However, SRC classifier has two drawbacks. Firstly, SRC classifier could not directly apply for object recognition [24]-[27] and face gesture. Secondly, the SRC classifier acquires large computational complexity. Thus, we need a low computation classifier, which can be used for face and object recognition,

In this paper, the polyhedron volume ratio classification (PVRC) method is proposed for image recognition [27]. With the minimum distance metric, the LRC and CM classifiers test the distance between the test sample vector and each class subspace, then the closest distance is identified as the best matched class. With the similar class-based classification, the PVRC classifier tests the ratio of the polyhedron volume enclosed by the test sample plus the class prototype samples over the polyhedron volume enclosed by the class prototype samples. The objective of LRC and PVRC are conceptually same, however, the difference between PVRC and LRC can be found in Section V-B. With Cayley-Menger matrix [21], in this paper, we first define the polyhedron volume and propose a new metric, called polyhedron volume ratio for inclusive and exclusive test sample to a specific class-subspace, for image classification. Based on mathematical (linear algebra) derivations, the PVRC is separated into training phase and test phase (TP&TP), where the computational cost in the test phase of PVRC is very small. In the test phase, the computation of the PVRC is almost equal to that of NN classifier and is much less than those of SRC, CRC and LRC.

The rest of the paper is organized as the follows. First, we review several well-known image classifiers in Section II. The computation of polyhedron volume and the PVRC classifier will be introduced in Section III. To reduce the computation, the fast PVRC classification procedure is described in Section IV. The analyses of the proposed PVRC are discussed in Section V. In Section VI, a number of experiments to show the effectiveness of the proposed classifier are present. The conclusions and future work are finally addressed in Section VII.

## II. REVIEW OF CLASS-SPECIFIC CLASSIFICATIONS

Let $Y = \{y_i^c, i=1,2,...,N_c, c=1,2,...,M\} \in R$ denote the prototype image set, where $y_i^c$ denotes the $i^{th}$ prototype image of the $c^{th}$ class, $M$ denotes the number of classes, and $N_c$ is the number of prototypes belonging to the $c^{th}$ class.

### II-A Linear Regression Classification Algorithm

For linear regression, each $a \times b$ image is transformed to vector by column concatenation as $y_i^c \in R^{a \times b} \to x_i^c \in R^{q \times 1}$, where $q = a\,b$. By using the concept that data vectors from the same class lie on a linear subspace, the LRC develops a class-specific model $X_c$ by stacking the $q$-dimensional image vectors as

$$X_c = [x_1^c \quad x_2^c \quad ... \quad x_{N_c}^c] \in R^{q \times N_c}. \quad (1)$$

Let $y$ be an unlabeled $a \times b$ test image and our problem is to classify $y$ as one of the classes. Similarly, we can transform $y$ into the vector form $x \in R^{q \times 1}$. If $y$ belongs to the $c^{th}$ class, it shall be well represented as a linear combination of the training images of the class as

$$x \approx x^c = X_c (X_c^T X_c)^{-1} X_c^T x. \quad (2)$$

If $X_c^T X_c$ is singular in (2), $X_c^T X_c$ will be replaced by $X_c^T X_c + 0.01I$, where $I$ is a unit matrix. In (2), the predicted vector $x^c$ can be treated as the projection of $x$ onto the $c^{th}$ class subspace. The LRC now calculates the distance measure between the predicted response vector $x^c$ and the original response vector $x$ as

$$d_c(x) = ||x - x^c||, \quad (3)$$

where $||*||$ means $L_2$-norm. The LRC classification rule is in favor of the class with the minimum distance of the class as

$$\min_{c^*} d_c(x), c=1,2,...,M. \quad (4)$$

### II-B. Collaborative representation based classification (CRC)

Suppose that we have $M$ classes of subjects, we can collect all class-specific models $X_c$, $c = 1, 2, …, M$, defined in (1) to form the complete data model as

$$X = [X_1 \quad X_2 \quad ... \quad X_M] \in R^{q \times MN_c}. \quad (5)$$

If the vector of all-class weighing parameters is denoted as $\beta \in R^{MN_c \times 1}$, it can be calculated as follows.

$$\beta = (X^T X)^{-1} X^T x. \quad (6)$$

For (6), if $X^T X$ is singular, $X^T X$ will be replaced by $X^T X + 0.01I$. The regularized residual of the $c^{th}$ class, $r^c$ is given as

$$r^c = \frac{||x - X_c \beta^c||_2}{||\beta^c||_2}, \quad (7)$$

where $\beta^c$ corresponding to the coefficient of the sample of class $c$ is a sectioned column of $\beta$. And the CRC classification rule in favor of the class with the minimum distance becomes

$$\min_{c^*} r^c, c=1,2,...,M. \quad (8)$$

### II-C Sparse representation based classification (SRC)

For the sparse representation-based classification (SRC), we first normalize the columns of $X$ stated into (5) to have unit $L_2$-norm. We can solve the $L_1$-norm minimization problem as:

$$\hat{g} = \arg\min_g ||g||_1 \quad \text{subject to } Xg = x. \quad (9)$$

Compute the regularized residuals $r^c$ as,

$$r^c = ||x - X_c \hat{g}^c||, \quad (10)$$

the SRC classification rule in favor of the class with the minimum distance can be expressed by

$$\min_{c^*} r^c, c=1,2,...,M. \quad (11)$$

## III. POLYHEDRON VOLUME RATIO CLASSIFICATION

Before the introduction of the polyhedron volume ratio classification (PVRC), we first review the concept of polyhedron volume of $n$ sample points. In [21], Cayley-Menger matrix constructed by $n$ ($n \geq 2$) sample points is given by:

$$Q_n^c = \begin{bmatrix} 0 & (b_{12}^c)^2 & (b_{13}^c)^2 & ... & (b_{1n}^c)^2 & 1 \\ (b_{21}^c)^2 & 0 & (b_{23}^c)^2 & ... & (b_{2n}^c)^2 & 1 \\ (b_{31}^c)^2 & (b_{32}^c)^2 & 0 & ... & (b_{3n}^c)^2 & 1 \\ ... & ... & ... & ... & ... & ... \\ (b_{n1}^c)^2 & (b_{n2}^c)^2 & (b_{n3}^c)^2 & ... & 0 & 1 \\ 1 & 1 & 1 & ... & 1 & 0 \end{bmatrix}, \quad (12)$$

where $b_{ij}^c$ denotes Euclidian distance of $x_i^c$ and $x_j^c$ as

$$b_{ij}^c = b_{ji}^c = ||x_i^c - x_j^c||. \quad (13)$$

The squared polyhedron volume of $n$ sample points can be defined as the determinant of Cayley-Menger matrix as,

$$v_n^2 = c_n \det(Q_n^c), \text{ for } n \geq 2, \quad (14)$$

with

$$c_n = \frac{(-1)^n}{2^{n-1}((n-1)!)^2}, \quad (15)$$

where det(*) denotes the determinant of the argument and $c_n$ represents the unifying factor in the computation of polyhedron volume.

To verify (14), we can check the following interesting examples. For $n = 2$, the Cayley-Menger matrix in (12) for two sample points, i.e., $x_1^c$ and $x_2^c$, is given as

$$Q_n^c = \begin{bmatrix} 0 & (b_{12}^c)^2 & 1 \\ (b_{21}^c)^2 & 0 & 1 \\ 1 & 1 & 0 \end{bmatrix}. \quad (16)$$

The determinant of $Q_2$ is $|Q_2^c| = 2(b_{12}^c)^2$ and $c_2 = 1/2$ with $b_{12}^c = \|x_1^c - x_2^c\|$. It is interested that the volume of 2-point polyhedron is the true distance between these two points.

According to Heron's formula [22], the true area of triangle of three points, $x_1^c$, $x_2^c$, and $x_3^c$, can be expressed by

$$S_{\Delta 123}^2 = [(b_{12}^c)^4 + (b_{13}^c)^4 + (b_{23}^c)^4 + 2(b_{12}^c)^2(b_{13}^c)^2(b_{23}^c)^2 \\ - 2(b_{12}^c)^2(b_{13}^c)^2 - 2(b_{12}^c)^2(b_{23}^c)^2 - 2(b_{13}^c)^2(b_{23}^c)^2]/16 \quad (17)$$

Similarly, the Cayley-Menger matrix of three points, $x_1^c$, $x_2^c$, and $x_3^c$ is given as

$$Q_3^c = \begin{bmatrix} 0 & (b_{12}^c)^2 & (b_{13}^c)^2 & 1 \\ (b_{21}^c)^2 & 0 & (b_{23}^c)^2 & 1 \\ (b_{31}^c)^2 & (b_{32}^c)^2 & 0 & 1 \\ 1 & 1 & 1 & 0 \end{bmatrix}. \quad (18)$$

The determinant of $Q_3^c$ stated in (18) is given as

$$|Q_3^c| = -[(b_{12}^c)^4 + (b_{13}^c)^4 + (b_{23}^c)^4 + 2(b_{12}^c)^2(b_{13}^c)^2(b_{23}^c)^2 \\ - 2(b_{12}^c)^2(b_{13}^c)^2 - 2(b_{12}^c)^2(b_{23}^c)^2 - 2(b_{13}^c)^2(b_{23}^c)^2]. \quad (19)$$

From (15), $c_3 = -1/16$ and (19), the polyhedron volume defined in (14) is equal to the true area of triangle enclosed by three points, $x_1^c$, $x_2^c$, and $x_3^c$ stated in (17).

Similarly, we can find 4-point Cayley-Menger matrix as

$$Q_4 = \begin{bmatrix} 0 & (b_{12}^c)^2 & (b_{13}^c)^2 & (b_{14}^c)^2 & 1 \\ (b_{21}^c)^2 & 0 & (b_{23}^c)^2 & (b_{24}^c)^2 & 1 \\ (b_{31}^c)^2 & (b_{32}^c)^2 & 0 & (b_{34}^c)^2 & 1 \\ (b_{41}^c)^2 & (b_{42}^c)^2 & (b_{43}^c)^2 & 0 & 1 \\ 1 & 1 & 1 & 1 & 0 \end{bmatrix}. \quad (20)$$

With $c_4=1/288$, the square volume of tetrahedron becomes $v_4^2 = \det(Q_4)/288$.

In [25], the computation of $n$-point polyhedron volume can be interpreted as the iterative computation of the shortest distance to the $(n-1)$-point surface while we treat $(n-1)$-point polyhedron volume as the base volume. The multiplication of the shortest distance to the base $(n-1)$-point volume will be the $n$-point polyhedron volume. If the number of sample data points are less than the dimension of data vectors, it can be easily inferred that the ratio of the $n$-point (with one test sample) polyhedron volume over the $(n-1)$-point polyhedron volume (with samples in a class) will be the perpendicular distance of the test sample to the $(n-1)$-point polyhedron. Based on the above interpretation of $n$-point and $(n-1)$-point polyhedron volumes, we introduce a new classifier, called polyhedron volume ratio classification (PVRC) method.

For images with $a \times b$ pixels, they are originally represented as $y_i^c \in R^{a \times b}$ by labeling the $c^{th}$ class while $y$ is an unlabeled test image. In the training phase, these $(n-1)$ prototype images are transformed to prototype vectors by concatenation of columns as $y_i^c \in R^{a \times b} \to x_i^c \in R^{q \times 1}$, with $q = a\,b$ for $i = 2, 3, \ldots, n$. Thus, each class contains $(n-1)$ prototype samples, which is $\{x_2^c, x_3^c, \cdots, x_n^c\} \in R^{q \times 1}$ for $c = 1, 2, \ldots, M$. The test image $y$ is also transferred to a test vector $x$ and renamed as $x = x_1 \in R^{q \times 1}$.

Thus, the metric of PVRC can be computed as

$$\rho^c = \frac{v_n^c}{v_{n-1}^c} = \frac{[c_n \det(Q_n^c)]^{1/2}}{[c_{n-1} \det(Q_{n-1}^c)]^{1/2}}, \quad (21)$$

where $v_n^c$ denotes the polyhedron volume of the test sample and $(n-1)$ prototype samples of the $c$-class and $v_{n-1}^c$ represents the polyhedron volume of $(n-1)$ prototype samples of the $c$-class. The decision rule of PVRC is simply to find the least ratio as

$$c^* = \min \rho^c, c = 1, 2, \ldots, M. \quad (22)$$

The above minimum polyhedron volume ratio means the least perpendicular distance between the test sample and the $c^*$-class. However, the computation of direct PVRC method involving the calculation of two determinates is too high for practical applications. Thus, a fast computation method for the PVRC is needed.

IV. FAST COMPUTATION METHOD AND DETAILED CLASSIFICATION PROCEDURE OF PVRC

The PVRC classifier computes the ratio of two polyhedron volumes, where the first polyhedron is enclosed by the test sample with $(n-1)$ prototype images in each class and the second polyhedron is enclosed by the $(n-1)$ prototype images in each class. To reduce the computation of the PVRC method, we should first explore the matrix segmentation and its determinant computation in **Theorem** 1.

***Theorem 1**: Matrix Segmentation and Its Determinant Computation*

Let $P$ be a square $(m+n) \times (m+n)$ matrix to be segmented into four submatrices, $A$, $B$, $C$ and $D$ are with the sizes of $m \times m$, $m \times n$, $n \times m$ and $n \times n$, respectively. If $D$ is a reversible square matrix, we can have the equality as,

$$\det(P) = \det\left(\begin{bmatrix} A & B \\ C & D \end{bmatrix}\right) = \det(A - BD^{-1}C) \times \det(D), \quad (23)$$

where $P = \begin{bmatrix} A & B \\ C & D \end{bmatrix}$.

The detailed proof of **Theorem** 1 can be found in Appendix A.

To further explore the relationship of the first and second polyhedron volumes, we can first divide $Q_n$ into four sub matrices as

$$Q_n^c = \begin{bmatrix} 0 & B_c \\ B_c^T & Q_{n-1}^c \end{bmatrix}, \quad (24)$$

where

$$Q_{n-1}^c = \begin{bmatrix} 0 & (b_{23}^c)^2 & \cdots & (b_{2n}^c)^2 & 1 \\ (b_{32}^c)^2 & 0 & \cdots & (b_{3n}^c)^2 & 1 \\ \cdots & \cdots & \cdots & \cdots & \cdots \\ (b_{n2}^c)^2 & (b_{n3}^c)^2 & \cdots & 0 & 1 \\ 1 & 1 & \cdots & 1 & 0 \end{bmatrix}, \quad (25)$$

and

$$B_c = \begin{bmatrix} (b_{12}^c)^2 & (b_{13}^c)^2 & \cdots & (b_{1n}^c)^2 & 1 \end{bmatrix}, \quad (26)$$

From (23), we know that the computation of determinant of $Q_n$ becomes as

$$\det(Q_n^c) = \det(Q_{n-1}^c)\det(-B_c(Q_{n-1}^c)^{-1}B_c^T). \quad (27)$$

By substituting (27) into (21), the computation of the PVRC metric can be simply expressed by

$$\rho^c = \frac{v_n^c}{v_{n-1}^c} = 2(n-1)^2 \left|\det(B_c(Q_{n-1}^c)^{-1}B_c^T)\right|^{1/2}. \quad (28)$$

It is noted that $B_c(Q_{n-1}^c)^{-1}B_c^T$ is a scalar, the determinant operation will not change it. Once we pre-compute $(Q_{n-1}^c)^{-1}$ in advance, the computation of ratio of two polyhedron volumes stated in (28) is less than any of the existing subspace-based classifications.

Finally, the classification procedure of the PVRC using fast computation method can be addressed as follows.

PVRC Training Phase:

By using (25), in the PVRC training phase, we should use all $(n-1)$ prototype samples, $\{x_2^c, x_3^c, \cdots, x_n^c\}$ to compute and store $(Q_{n-1}^c)^{-1}$ for $c=1, 2, \ldots, M$;

PVRC Testing Phase:

For any test sample, $x_1$, we should use all prototype samples $\{x_2^c, x_3^c, \cdots, x_n^c\} \in R^{q \times 1}$ for $c=1, 2, \ldots, M$ to
1. compute all distances of the test sample to all prototype samples as:
   $b_{1i}^c = ||x_1 - x_i^c||$, for $i=2, 3, \ldots, n$ and $c=1, 2, \ldots, M$;
2. construct $B_c$ by using (26) for $c=1, 2, \ldots, M$;
3. compute absolute value of square volume ratio as

$$\xi_c = \left|\det(B_c(Q_{n-1}^c)^{-1}B_c^T)\right|; \quad (29)$$

4. pick $c^* = \min_{c^*} \xi_c, c = 1, 2, \ldots, M$.

Actually, for true polyhedron volume, we need to include unifying factor $c_n$ and square root operation as (25). In Step 3, the absolute value of square volume ratio with the factor related $c_n$ is intentionally ignored due to fixed $n$. Besides, the square root operation is also skipped without affecting the minimization process. When the numbers of samples in the classes are various, we should further include a factor of $4(n-1)^4$ in (29) for correct classification.

It is noted that the prototype samples are very similar or the number of prototype samples is less than the dimension of sample vectors, the linear regression solutions stated in (2), (6), (9) might not be stable due to the possible singular problem in matrix inversion. To solve the singular problems, we can add a small identity matrix for most linear regression problem. For the PVRC computation, involving either (25) or (27), the singularity of matrix inversion only exists for similar prototype samples. Without scarifying the detection performance, the addition of a small identify matrix to $Q_{n-1}^c$ is also possible.

V. ANALYSIS OF PVRC CLASSIFIER

In this section, we analyze the classification rule of PVRC classifier firstly. Then, the interpretation among the PVRC, CM, NN, LRC, CRC, TPTSSR and SRC is introduced. In the last, the computational cost of PVRC is described.

V-A. Classification rule of PVRC

From (28) and (29), we know that the decision rule of the PVRC is based on two matrixes $B_c$ and $(Q_{n-1}^c)^{-1}$. The matrix $(Q_{n-1}^c)^{-1}$ is unrelated to the test sample while the matrix $B_c$ is formed by the distances between the test sample and the prototype samples of each class subspace. So, we learn that the classification rule of the PVRC can be seen as the combination of the distances between the test sample and the samples of each class subspace. That is to say, the PVRC classifier can be treated as the extension of the NN classifier. However, the combination distance metric of PVRC classifier is superior to the single distance metric of NN classifier.

V-B. Computation of PVRC

The most computational cost of the PVRC classifier will be in the training phase by constructing $Q_{n-1}^c$ and computing its matrix inversion, $(Q_{n-1}^c)^{-1}$ for all classes. The computational complexity will be in the order of $O(n^3)$ for each class. The computational cost in the testing phase of the PVRC classifier is very small. The detailed computational complexity of the testing phase of the PVRC classifier is discussed as follows. From (29), the computation in the testing phase mainly contains two parts. The first part is the computation cost in construction of the vector, $B_c$ stated in (26). If dimensionality of the sample vector is with $q$, Its computational complexity is about in the order of $O(nq)$ for each class, The total computation complexity will be in the order of $O(nq)+O(n^2)$. Since the size of sample vectors, $q$ is much greater than the number of sample vectors, $n$, Thus, we can say that the computational complexity of the PVRC classifier is approximately equal to in the order of $O(nq)$, i.e., the cost in construction of the vector $B_c$. It is easy to know that the computation in constituting the matrix $B_c$ is approximately equal to the cost of the NN classifier. In summary, the computation cost of the PVRC is approximately equal to the cost of NN classifier, which is much less than the costs of the SRC, CRC, LRC and TPTSSR classifiers. The detailed advantages in various classification performances of the PVRC classifier will be verified by experimental results.

*V-C Interpretation of well-known classifiers*

**Fig. 1** also shows the concept of the other classification methods based on class linear regression and class mean. The NN classifier uses the distance between the test sample and each prototype sample to classify the test sample. The CM classifier uses the distance between the test sample and the mean sample of each class to classify the test sample. The other classifiers such as the CRC, TPTSSR and SRC classifiers try to find the better approximations of the subspace of the class to achieve better recognition performances. Though the NN, CM, LRC, CRC, TPTSSR and SRC classifier are different, they have one thing in common, which is that they all need to compute the distance between the test sample and the approximated class center or the subspace, which is derived from the prototype samples or the re-sampled prototype samples, for classifications. The performances of the NN, CM, LRC, CRC, TPTSSR and SRC classifiers only depend on the closeness of their approximated class centers or subspaces.

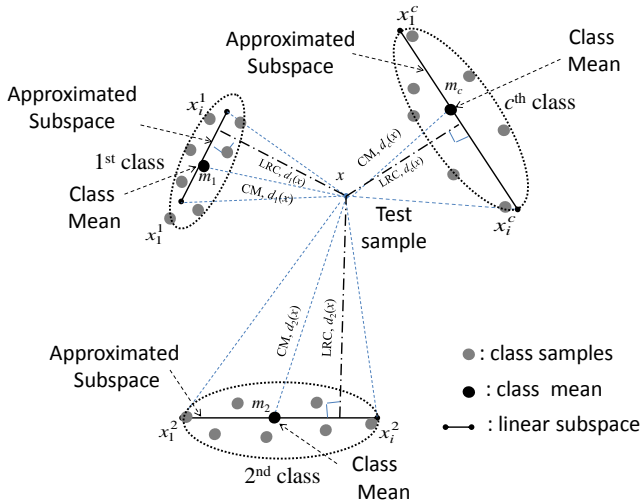

**Fig. 1.** Conceptual relationships among the proposed PVRC and the projection-based classification methods

*V-D. Comparisons of PVRC and LRC classifiers*

The ratio of polyhedron volumes with/without extra the test sample to the class samples can be interpreted as the computation of the perpendicular distance to the prototype samples. From **Fig. 1**, the PVRC find the minimum distance from the polyhedron of the enclosed class data vectors while the LRC find the least distance from spanned subspace formed by class data vectors, The differences between the LRC and the PVRC are addressed as follows. From computation point of view, the PVRC is different from the LRC, where the former is derived through a mathematical (linear algebra) approach as stated in (21) while the latter is obtained from a statistical way to minimize the statistical predictin error as addressed in (2).

By simulations, **Fig. 2** shows the ratio of the computed distances obtained by the PVRC and the LRC if they use the same randomly-generated data points. To compute the ratio between PVRC's distance and LRC's distance, the corresponding MATLAB code can be downloaded from [http://pan.baidu.com/s/1pJx9Z0B]. The ratio between the PVRC's distance and LRC's distance is variation from 5 to 9. Therefore, we learnt that the results of PVRC and LPC are different.

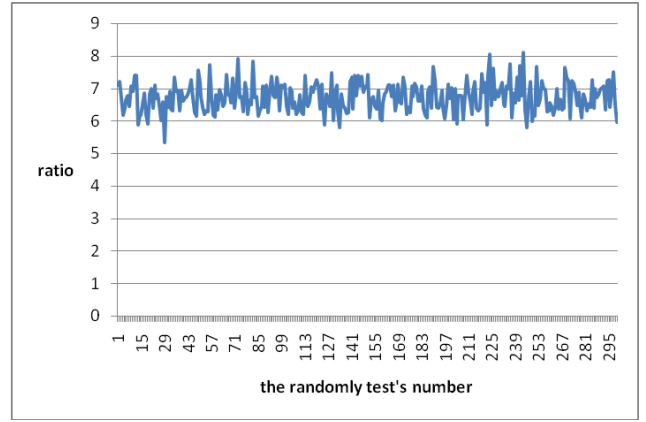

**Fig. 2.** The ratio of PVRC's and LRC's distances for 300 tests.

For computation analyses, if we separate the LRC into training phase and test phase (TP&TP) such that we could achieve a faster classification. In the test phase, the computational complexity of the LRC is $O(Q^2)$ while the PVRC and NN is about $O(QN_c)$. Thus, the LRC acquires much more computation than the PVRC and NN classifiers since $Q \gg N_c$. The similar results also verified from **code1** [http://pan.baidu.com/s/1pJx9Z0B] that the computational time of PVRC is almost equal to that of the NN, which is much simpler than the original LRC and TP&TP LRC. Besides, the LRC needs about $O(MQQ)$ storage space to store pre-computed results, while the PVRC needs about $O(MN_cN_c)$ storage space.

## VI. EXPERIMENTAL RESULTS

The classification performances of the PVRC classifier are compared to those of the LRC, SRC, CRC, TPTSSR, CM and NN classifiers. Two test schemes are taken for comparisons. Firstly "leave one out" scheme: All images within prototype database are taken as the test samples. When an image is used as the test sample, it is not used as a prototype and it is removed from the prototype set during the classifications; Secondly "First $N$" scheme: the first $N$ face images of each class are used as the prototype set. The rest face images of test database are used as test samples. The recognition rate (RR) is used to evaluate the performance of new algorithms.

All experiments are implemented using the MATLAB R2009a under Intel (R) Core (TM) i5 CPU 760 with a clock speed of 2.80GHz, 2.80GHz and 3GB RAM.

*VI-A Computational Cost*

The computational cost of the classification procedure of several classifier is depended on the number of samples, is not depended on the specific value of samples. So, we only need provide the result of one database. The other databases are similar. In **Table I**, the computational cost of each query of several classifiers on coil100 object database [23] is described, which is corresponding to the run time of the second experiments in the part B. From **Table I**, after computer MATLAB simulations, we learn that the computational costs (unit: seconds) of the PVRC, NN and CM classifiers are similar, which are less than those of the CRC and LRC classifiers and much less than that of the SRC classifier. It is noted that the computational cost of the first phase of the TPTSSR is similar to that of CRC, the computational cost of the second phase of the TPTSSR is according to the number of samples chosen in the first phase. So, the total computational cost of the TPTSSR is more than that of the CRC, which is also more than that of the proposed PVRC.

**Table I:** COMPUTATIONAL COST OF EACH QUERY ON COIL OBJECT DATABASE USING THE "FIRST N" SCHEME (UNIT: SECONDS)

| Classifier | RR (4) | RR (8) |
|---|---|---|
| NN | 0.00443 | 0.00850 |
| CM | 0.00742 | 0.01272 |
| LRC | 0.05696 | 0.09999 |
| CRC | 0.29658 | 1.18030 |
| SRC | 3.06740 | 7.18340 |
| PVRC | 0.00546 | 0.00903 |

*VI-B Object recognition on coil object database*

The Coil-100 data set [23] was widely used as an object-recognition benchmark [24]-[27]. In this data set, there are 100 objects and each object has 72 different views (images) that are taken every 5° around an axis passing through the object. Each image is a 128×128 color one with R, G, B channels. We use only a limited number of views per objects for experiments. In our experiments, 12 different views per object (0°, 30°, 60°, 90°, 120°, 150°, 180°, 210°, 240°, 270°, 300° and 330°) were used, shown in **Fig. 3**. So the subset of Coil-100 data set contains 1200 images, and all images in subset of Coil-100 database were manually cropped into a 32×32 color one with R, G, B channels.

In the first experiment, we adopt the "leave one out" scheme on coil object database. The results are listed in **Table II**. In the second experiment, we test the recognition rate on coil object database using "first *N*" scheme. The results are described in **Fig. 4** and **Table III**.

In **Table II**, the recognition rate (RR) of PVRC classifier outperforms the RRs of the SRC, LRC, CRC, NN and CM classifiers with 7.08%, 1.50%, 13.25%, 6.75%, and 15.83% improvements, respectively. In **Table III**, the RR of the PVRC classifier outperforms the RRs of the SRC, LRC, CRC, NN and CM classifiers, with 7.00%, 1.00%, 30.00%, 4.88%, and 4.00% improvements, respectively while the first 4 samples are used as prototype. The RR of the PVRC classifier outperforms the RRs of them with 4.72%, 1.95%, 9.25%, 6.45%, and 13.25% improvements, respectively while the first 8 samples are used as the prototypes. From **Fig.4**, we can know that the RR of the PVRC classifier outperforms the best RR of the TPTSSR with 10.12% (8.50%) improvement when the first 4 (8) samples of each class are used as the prototypes. The horizontal axis of **Fig. 4** is the *number of* nearest neighbors of the TPTSSR.

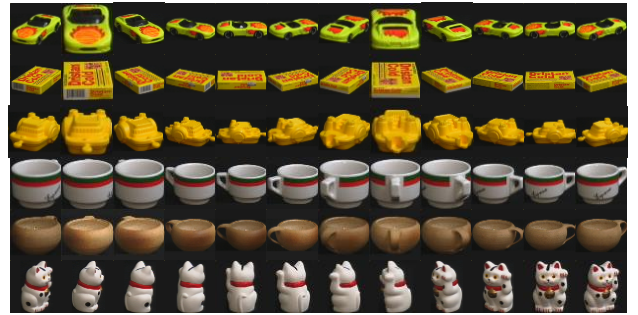

**Fig. 3.** Some images of the subset of Coil-100 database

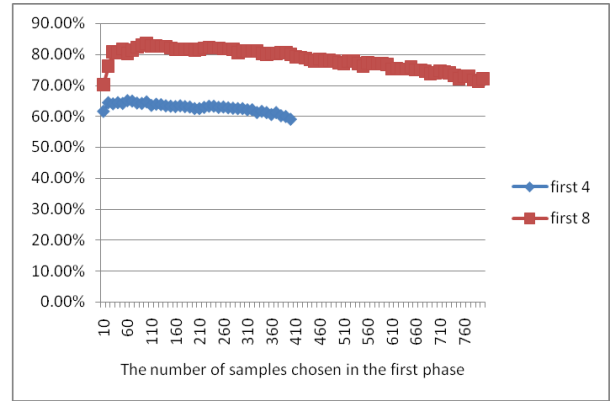

**Fig. 4.** The RR of the TPTSSR classifier on coil object database using the "first *N*" scheme

**Table II:** RECOGNITION RATES (RRs) OF SEVERAL CLASSIFIERS ON COIL OBJECT DATABASE USING THE "LEAVE ONE OUT" SCHEME

| Classifier | NN | CM | LRC |
|---|---|---|---|
| RR | 86.25% | 77.17% | 91.50% |
| Classifier | CRC | SRC | PVRC |
| RR | 79.75% | 85.92% | 93.00% |

**Table III:** RECOGNITION RATES (RRs) OF SEVERAL CLASSIFIERS ON COIL OBJECT DATABASE USING THE "FIRST N" SCHEME

| Classifier | RR (4) | RR (8) |
|---|---|---|
| NN | 69.87% | 85.55% |
| CM | 70.75% | 78.75% |
| LRC | 73.75% | 90.05% |
| CRC | 65.00% | 82.75% |
| SRC | 67.75% | 87.25% |
| PVRC | 74.75% | 92.00% |

*VI-C Object recognition on eth80 object database*

In the eth80-cropped-close128 object database [28] [29], all images are cropped, so that they contain only the object without any border area. In addition, they are rescaled to a size of 128×128 pixels. Again, the scale is left the same for all images of the same object. This dataset is useful when no derivatives are needed. In our experimental, all images are resized to 32×32 gray images. Some images of eth80 object database are shown in **Fig. 5**.

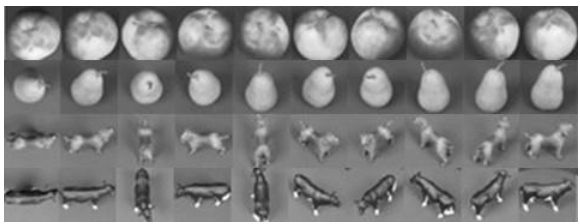

**Fig. 5.** Some sampled images of eth80 database

In the third experiment, we adopt the "leave one out" scheme on eth80 object database. The result is listed in **Table IV**. In the fourth experiment, we test the recognition rate on eth80 object database using "first *N*" scheme. The results are described in **Fig. 6** and **Table V**.

In **Table IV**, the recognition rate (RR) of the PVRC classifier outperforms the RRs of the SRC, LRC, CRC, NN and CM classifiers with 11.02%, 4.75%, 24.45%, 11.25%, and 29.05% improvements, respectively. In **Table V**, the RR of the PVRC classifier outperforms the RRs of the SRC, LRC, CRC, NN and CM classifiers with 7.56%, 4.22%, 9.73%, 3.07%, and 5.10% improvements, respectively while the first 4 samples are used as prototype. The RR of the PVRC classifier outperforms the RRs of them with 6.15%, 4.40%, 11.24%, 3.5%, and 6.25% improvements, respectively while the first 6 samples are used as prototype. From **Fig. 6**, we learnt that the RR of PVRC classifier outperforms the best RR of the TPTSSR 12.03% (13.86%) when the first 4 (6) samples of each class are used as prototype. The horizontal axis of **Fig. 6** is the *number of* nearest neighbors of the TPTSSR.

**Table IV:** THE RR OF SEVERAL CLASSIFIERS ON ETH80 OBJECT DATABASE USING THE "LEAVE ONE OUT" SCHEME

| Classifier | NN | CM | LRC |
|---|---|---|---|
| RR | 64.60% | 46.80% | 71.10% |
| Classifier | CRC | SRC | PVRC |
| RR | 51.40% | 64.82% | 75.85% |

**Table V:** RECOGNITION RATES (RRs) OF SEVERAL CLASSIFIERS ON ETH80 OBJECT DATABASE USING THE "FIRST N" SCHEME

| Classifier | RR (4) | RR (6) |
|---|---|---|
| NN | 19.46% | 21.86% |
| CM | 17.43% | 19.11% |
| LRC | 18.31% | 20.96% |
| CRC | 12.80% | 14.11% |
| SRC | 14.97% | 19.21% |
| PVRC | 22.53% | 25.36% |

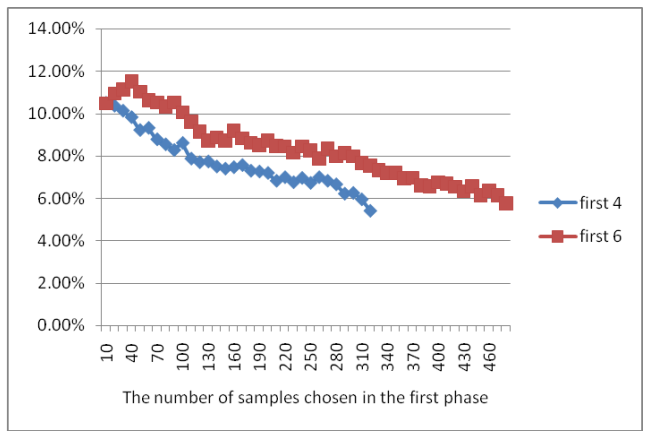

**Fig. 6.** The RR of TPTSSR classifier on eth80 object database using the "first *N*" scheme

*VI-D Object recognition on soil47 object database*

The Soil-47 data set [30] was widely used as an object-recognition benchmark. In the data set, there are 47 objects and each object has 21 different views (images) that are taken every 9° around an axis passing through the object. In the experimental, the subset of soil-47 includes 966 images of 46 objects. Each color image is downsampled to a 24 × 32 gray image. **Fig**. 7 shows some sampled images of these objects.

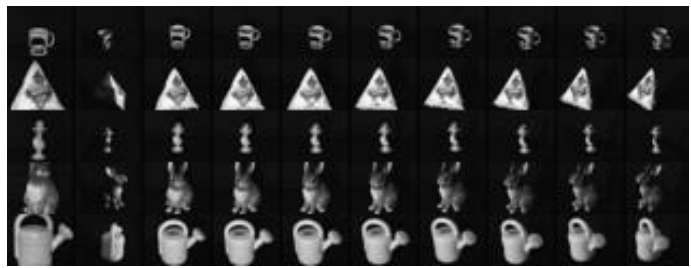

**Fig. 7** Some sampled images of SOIL-47 data set

In the fifth experiment, we adopt the "leave one out" scheme on eth80 object database. The result is listed in **Table VI**. In the sixth experiment, we test the recognition rate on eth80 object database using "first *N*" scheme. The results are described in **Fig. 8** and **Table VII**.

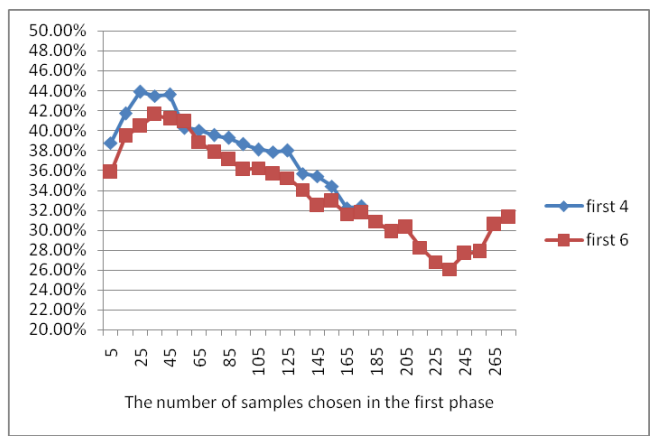

**Fig. 8.** The RR of TPTSSR classifier on Soil47 object database using the "first *N*" scheme

**Table VI**: THE RR OF SEVERAL CLASSIFIERS ON SOIL47 OBJECT DATABASE USING THE "LEAVE ONE OUT" SCHEME

| Classifier | NN | CM | LRC |
|---|---|---|---|
| RR | 64.60% | 44.62% | 72.67% |
| Classifier | CRC | SRC | PVRC |
| RR | 64.18% | 71.01% | 75.85% |

**Table VII**: RECOGNITION RATE (RR) OF SEVERAL CLASSIFIERS ON SOIL47 OBJECT DATABASE USING THE "FIRST N" SCHEME

| Classifier | RR (4) | RR (6) |
|---|---|---|
| NN | 42.46% | 39.57% |
| CM | 39.39% | 35.07% |
| LRC | 49.74% | 50.15% |
| CRC | 45.01% | 43.04% |
| SRC | 48.47% | 48.70% |
| PVRC | 50.26% | 51.31% |

In **Table VI**, the recognition rate (RR) of the PVRC classifier outperforms the RRs of the SRC, LRC, CRC, NN and CM classifiers with 4.84%, 3.18%, 11.67%, 11.25%, and 31.23% improvement, respectively. In **Table VII**, the RR of the PVRC classifier outperforms the RRs of the SRC, LRC, CRC, NN and CM classifiers with 1.79%, 0.52%, 5.25%, 7.80%, and 10.87% improvements, respectively while the first 4 samples are used as prototype. The RR of the PVRC classifier outperforms the RR of them with 2.61%, 1.16%, 8.27%, 11.74%, and 16.24% improvements, respectively while the first 6 samples are used as prototype. From **Fig. 8**, we learnt that the RR of PVRC classifier outperforms the best RR of the TPTSSR with 7.45% (10.71%) improvements when the first 4 (6) samples of each class are used as the prototypes. The horizontal axis of **Fig. 8** is the *number of* nearest neighbors of the TPTSSR.

*VI-E Face recognition on GT face database*

Georgia Tech face database [31] contains images of 50 people taken in two or three sessions between 06/01/99 and 11/15/99 at the Center for Signal and Image Processing at Georgia Institute of Technology. All people in the database are represented by 15 color JPEG images with cluttered background taken at resolution 640×480 pixels. The average size of the faces in these images is 150×150 pixels. The pictures show frontal and/or tilted faces with different facial expressions, lighting conditions and scale. Each image is manually of the GT face database is manually cropped in a 30×40 gray image. **Fig. 9** shows some selected face images of GT face database.

In the seventh experiment, we adopt the "leave one out" scheme on GT face database. The results are listed in **Table VIII**. In the eighth experiment, we test the recognition rate on GT face database using "first *N*" scheme. The results are described in **Fig. 10** and **Table VIIII**.

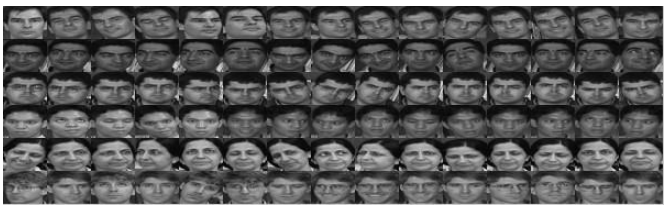

**Fig. 9.** Some selected face images of GT face database

**TABLE VIII**: THE RR OF SEVERAL CLASSIFIERS ON GT FACE DATABASE USING THE "LEAVE ONE OUT" SCHEME

| Classifier | NN | CM | LRC |
|---|---|---|---|
| RR | 82.64% | 73.60% | 84.67% |
| Classifier | CRC | SRC | PVRC |
| RR | 74.27% | 83.87% | 87.20% |

**Table VIIII**: THE RECOGNITION RATE (RR) OF SEVERAL CLASSIFIERS ON GT FACE DATABASE USING THE "FIRST N" SCHEME

| Classifier | RR (3) | RR (6) |
|---|---|---|
| NN | 49.83% | 67.11% |
| CM | 48.50% | 54.22% |
| LRC | 51.83% | 68.44% |
| CRC | 46.00% | 61.33% |
| SRC | 53.17% | 68.89% |
| PVRC | 53.83% | 72.89% |

In **Table VIII**, the recognition rate (RR) of the PVRC classifier outperforms the RR of the SRC, LRC, CRC, NN and CM classifiers with 3.33%, 2.53%, 12.93%, 4.56%, and 13.60% improvements, respectively. In **Table VIIII**, compared with SRC, LRC, CRC, NN and CM classifiers, the RR of PVRC classifier outperforms the RRs of them with 0.66%, 1.82%, 7.83%, 6.36%, and 6.18% improvements, respectively, when the first 3 samples are used as the prototypes. The RR of the PVRC classifier outperforms the RRs of them with 4.00%, 4.45%, 11.56%, 5.75%, and 18.67% improvements, respectively, when the first 6 samples are used as the prototypes. From **Fig.10**, we learn that the RR of the PVRC classifier outperforms the best RR of the TPTSSR with 0.83% (6.45%) improvement when the first 3 (6) samples of each class are used as prototype. The horizontal axis of **Fig. 10** is the *number of* nearest neighbors of the TPTSSR.

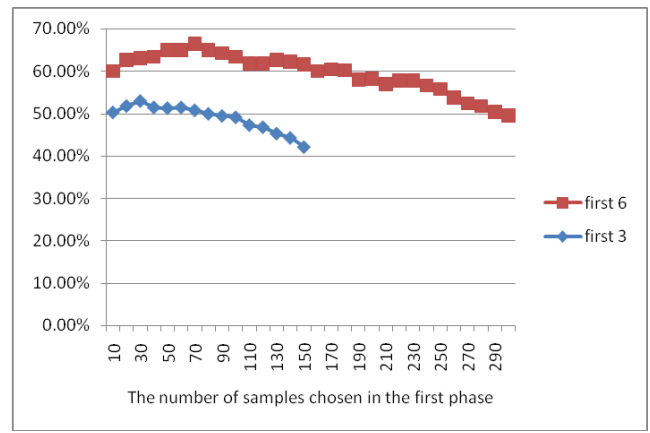

**Fig.10.** The RR of TPTSSR classifier on GT face database using the "first *N*" scheme

*VI-F Face recognition on UMIST face database*

The Sheffield (previously UMIST) Face Database [32] consists of 564 images of 20 individuals, which are mixed with race, gender, and appearance. Each individual is shown in a range of poses from profile to frontal views - each in a separate directory labelled 1a, 1b, ... 1t and images are numbered consecutively as they were taken. The files are all in PGM format, approximately 220×220 pixels with 256-bit grey-scale. In the experiment, a subset of UMIST face database contains 480 face images of 20 individuals. Each individual has 24 face images. All images of the subset are manually cropped in 40×50 gray images. **Fig. 11** exhibits some sampled face images of UMIST face database.

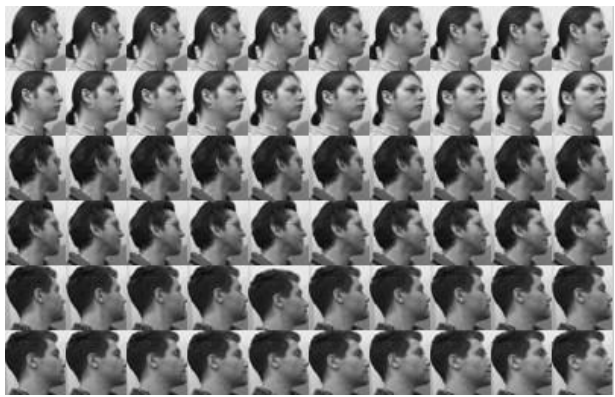

**Fig. 11.** Some sampled face images of UMIST face database

**Table VV**: The RR of Several Classifiers on UMIST Face Database Using the "Leave One Out" Scheme

| Classifier | NN | CM | LRC |
|---|---|---|---|
| RR | 99.25% | 95.75% | 99.75% |
| Classifier | CRC | SRC | PVRC |
| RR | 98.25% | 99.75% | 99.75% |

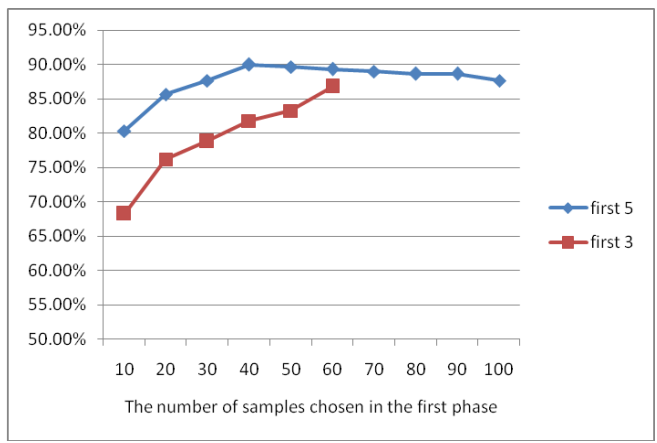

**Fig.12:** The RR of the TPTSSR classifier on UMIST face database using the "first *N*" scheme

In the ninth experiment, we adopt the "leave one out" scheme on UMIST face database. The results are listed in **Table VV**. In the tenth experiment, we test the recognition rate on GT face database using "first *N*" scheme. The results are described in **Fig. 12** and **Table VVI**.

**Table VVI :** Recognition Rate (RR) of Several Classifiers on UMIST Face Database Using the "First N" Scheme

| Classifier | RR (3) | RR (5) |
|---|---|---|
| NN | 90.00% | 89.67% |
| CM | 89.12% | 88.00% |
| LRC | 88.24% | 87.33% |
| CRC | 84.41% | 83.67% |
| SRC | 86.47% | 88.33% |
| PVRC | 90.00% | 90.00% |

In **Table VV**, the recognition rate (RR) of the PVRC classifier outperforms the RRs of the SRC, LRC, CRC, NN and CM classifiers, with 0.00%, 0.00%, 1.50%, 0.50%, 4.00% improvements, respectively. In **Table VVI**, the RR of PVRC classifier outperforms the RRs of the SRC, LRC, CRC, NN and CM classifiers with 3.53%, 1.76%, 5.59%, 0.00%, and 0.88%, improvements, respectively, when the first 3 samples are used as prototype. The RR of PVRC classifier outperforms the RRs of the SRC, LRC, CRC, NN and CM classifiers with 1.67%, 2.67%, 5.33%, 0.33%, and 2.00% improvements, respectively, when the first 5 samples are used as the prototypes. From **Fig.12**, we learnt that the RR of the PVRC classifier outperforms the best RR of the TPTSSR 3.24% (0.00%) when the first 3 (5) samples of each class are used as prototype. The horizontal axis of **Fig. 12** is the *number of* nearest neighbors of the TPTSSR.

## VII. CONCLUSION

In this paper, a novel classifier called polyhedron volume ratio classification (PVRC) is proposed for image recognition. The PVRC classifier uses the test sample vector and the original class prototype samples to calculate the distance between the test sample and the corresponding class prototype samples. The class-based distance computation is conceptually similar to the LRC and CM classifiers. However, the PVRC computes the distance between the test sample and the class subspace by the ratio of two polyhedron volumes instead of calculating the distance between the test sample and the simplified class features, such as the class means or approximated class subspaces. The proposed PVRC classifier, which computes the perpendicular distance of the test sample to the class polyhedron, achieves the better recognition rate than the NN, CM, CRC, SRC, TPTSSR and LRC classifiers. The computation of the PVRC in the testing phase is similar to those of the NN and CM classifiers. Thus, the PRVR takes less computation than the CRC, LRC and SRC classifiers. All experimental results confirm the effectiveness and analyses of the proposed classification algorithm.

**Appendix A: Matrix Segmentation and Its Determinant Computation**

Let $P = \begin{bmatrix} A & B \\ C & D \end{bmatrix}$ be a square matrix $(m+n) \times (m+n)$ to be segmented into four submatrices, *A*, *B*, *C* and *D* are with sizes of $m \times m$, $m \times n$, $n \times m$ and $n \times n$, respectively. It is noted that *D* should be a reversible square matrix. Then, we can prove:

$$\det(P) = \det\left(\begin{bmatrix} A & B \\ C & D \end{bmatrix}\right) = \det(A - BD^{-1}C) \times \det(D).$$

**Proof:**

Give a square matrix, $\begin{bmatrix} I_m & O \\ -D^{-1}C & I_n \end{bmatrix}$, where $I_m, I_n$ are $m \times m$, and $n \times n$ identify matrices, respectively, and $O$ is zeros matrix. Then, it is easy to verify

$$\begin{bmatrix} A & B \\ C & D \end{bmatrix}\begin{bmatrix} I_m & O \\ -D^{-1}C & I_n \end{bmatrix} = \begin{bmatrix} A - BD^{-1}C & B \\ O & D \end{bmatrix} \quad (A1)$$

By using the existing properties of determinant, $\det(AB) = \det(A)\det(B)$ and $\det\begin{bmatrix} E & F \\ O & G \end{bmatrix} = \det(E)\det(G)$, we can compute the determinant of both sides of (A1) to obtain,

$$\begin{aligned}
&\det(\begin{bmatrix} A & B \\ C & D \end{bmatrix})\det(\begin{bmatrix} I_m & O \\ -D^{-1}C & I_n \end{bmatrix}) \\
&= \det(\begin{bmatrix} A - BD^{-1}C & B \\ O & D \end{bmatrix}) \quad (A2) \\
&= \det(D)\det(A - BD^{-1}C)
\end{aligned}$$

In (A2), it is also straightforward to prove that $\det\begin{bmatrix} I_m & O \\ -D^{-1}C & I_n \end{bmatrix} = \det(I_m)\det(I_n) = 1$.

End of Proof.